\documentclass[10pt,twocolumn,letterpaper]{article}
\usepackage[pagenumbers]{mytemplate} % To force page numbers, e.g. for an arXiv version

\usepackage[dvipsnames]{xcolor}

\usepackage{float}
\usepackage{epsfig}
\usepackage{xcolor}
\usepackage{graphicx}
\usepackage{float}
\usepackage[utf8]{inputenc}
\usepackage{caption}
\usepackage{subcaption}
\usepackage{array}
\usepackage{algorithm}
\usepackage{algpseudocode}
\usepackage{multirow}
\usepackage{etoolbox}
\usepackage{anyfontsize}
\usepackage{booktabs}
\usepackage{colortbl}
\usepackage{bm}
\usepackage{pifont}
\usepackage{adjustbox}
\usepackage[accsupp]{axessibility}

\newcommand{\myparagraph}[1]{\noindent\textbf{#1}}

\newcommand{\cmark}{\textcolor{green}{\ding{51}}} % Green checkmark
\newcommand{\ccmark}{\cmark\kern-0.4em\cmark} % Green double checkmark
\newcolumntype{R}[2]{%
    >{\adjustbox{angle=#1,lap=\width-(#2)}\bgroup}%
    l%
    <{\egroup}%
}

\newtoggle{arXiv}

\definecolor{mytemplateblue}{rgb}{0.21,0.49,0.74}
\usepackage[pagebackref,breaklinks,colorlinks,citecolor=mytemplateblue]{hyperref}
\toggletrue{arXiv}

\title{RealityAvatar: Towards Realistic Loose Clothing Modeling \\ in Animatable 3D Gaussian Avatars}

\author{
Yahui Li\textsuperscript{1} \quad
Zhi Zeng\textsuperscript{1}\textsuperscript{\textdagger} \quad
Liming Pang\textsuperscript{1} \quad
Guixuan Zhang\textsuperscript{1} \quad 
Shuwu Zhang\textsuperscript{1} \\
\textsuperscript{1}Beijing University of Posts and Telecommunications \\
{\tt\small \{liyahui, zhi.zeng, plm, guixuan.zhang, shuwu.zhang\}@bupt.edu.cn}
    }

\begin{document}

\maketitle

\renewcommand{\thefootnote}{\textdagger}
\footnotetext[1]{Corresponding author}

\begin{abstract}
Modeling animatable human avatars from monocular or multi-view videos has been widely studied, with recent approaches leveraging neural radiance fields (NeRFs) or 3D Gaussian Splatting (3DGS) achieving impressive results in novel-view and novel-pose synthesis. However, existing methods often struggle to accurately capture the dynamics of loose clothing, as they primarily rely on global pose conditioning or static per-frame representations, leading to oversmoothing and temporal inconsistencies in non-rigid regions. To address this, We propose RealityAvatar, an efficient framework for high-fidelity digital human modeling, specifically targeting loosely dressed avatars. Our method leverages 3D Gaussian Splatting to capture complex clothing deformations and motion dynamics while ensuring geometric consistency. By incorporating a motion trend module and a latentbone encoder, we explicitly model pose-dependent deformations and temporal variations in clothing behavior. Extensive experiments on benchmark datasets demonstrate the effectiveness of our approach in capturing fine-grained clothing deformations and motion-driven shape variations. Our method significantly enhances structural fidelity and perceptual quality in dynamic human reconstruction, particularly in non-rigid regions, while achieving better consistency across temporal frames.

\textbf{KeyWords}: 3D Gaussian Splatting;  Animatable Avatar; Loose Clothing Dynamics; Temporal Modeling
\end{abstract}  

\section{Introduction}
\label{sec:intro}

In recent years, animatable digital humans have gained widespread attention as realistic representations of human characters in virtual worlds. Compared to traditional static or animation-driven virtual characters, drivable digital humans provide a more immersive interactive experience and are widely applied in areas such as virtual livestreaming, film production, and metaverse avatars. However, existing digital human modeling technologies still face challenges in terms of accuracy, adaptability to dynamic motion, and computational efficiency.

Animatable digital human modeling based on 3D observation data typically employs either implicit or explicit representations. explicit representations offer higher efficiency in modeling human surfaces~\cite{alldieck2018detailed,habermann2021real,zheng2023pointavatar}. Traditional explicit representaitons, like Mesh-based explicit methods, suffer from fixed topology, making it difficult to model fine details such as clothing wrinkles.
Digital human modeling based on 3D observation data typically employs either implicit or explicit representations. On the other hand, Implicit methods~\cite{mescheder2019occupancy,DeepSDF,xie2022neural}, such as NeRF-based deformable portrait modeling~\cite{mildenhall2020nerf}, utilize inverse skinning to map image observations from the posed space to the canonical space. However, implicit 3D volumetric representations struggle to efficiently represent human surfaces, and the inverse skinning process often introduces ambiguities that degrade fine-detail capture. 

Recently, 3D Gaussian Splatting (3DGS)~\cite{kerbl20233d} has emerged as an efficient alternative to NeRF, achieving remarkable progress in static scene modeling due to its superior inference speed and rendering quality. Methods such as D3GA~\cite{zielonka2023drivable} and GaussianAvatar~\cite{hu2024gaussianavatar} have leveraged 3DGS to enable the generation of animatable human models. These approaches combine linear blend skinning (LBS) for skeleton-driven motion, MLP-based non-rigid motion modeling, and standard radiance fields to animate digital humans based on pose control. However, these methods still face limitations when handling complex non-rigid deformations, such as loose clothing, making it challenging to maintain dynamic consistency.

Existing methods primarily rely on tight-fitting clothing datasets (e.g., ZJU-Mocap Dataset~\cite{peng2020neural}, ENeRF-Outdoor Dataset~\cite{lin2022efficient}) for modeling, whereas real-world digital human modeling inevitably involves large-scale motion and variations in loose clothing. For example, when a person abruptly stops after rotating, their pose may remain unchanged, but their clothing naturally drapes in a dynamically varying manner. Relying solely on static pose inputs fails to sufficiently model such dynamic non-rigid deformations, leading to artifacts and distortions.

To address these issues, we propose \textbf{RealityAvatar}, which leverages the power of 3D Gaussian Splatting to model digital humans wearing loose clothing in dynamic motion. We decompose the transformation from canonical space to observation space into two parts: a motion trend module (to capture the overall motion trend) and a body transformation module. In modeling human shape variations, we introduce a novel skeletal feature encoding method and adopt implicit encoding to enhance the generalization ability of clothing deformations. In addition, we incorporate recurrent neural networks (RNNs) to capture the dynamic characteristics of clothing affected by past motion states, enabling more accurate descriptions of complex deformations. In addition, we use a lightweight MLP for color decoding, which is specifically designed to adapt to dynamic lighting changes, resulting in more realistic appearance rendering.

Experimental results demonstrate that our approach, under fast training, accurately models dynamic clothing deformations and outperforms existing methods in terms of rendering quality and computational efficiency. While our training speed is slightly slower than methods focusing purely on fast training, RealityAvatar enables pose-dependent non-rigid deformation modeling, producing higher-quality rendering results. Compared to other approaches specialized in dynamic clothing modeling~\cite{chen2024within,lei2023gart}, our method significantly improves training efficiency while maintaining superior performance. In summary, our main contributions are as follows: 
\begin{itemize}
    \item We innovatively employ recurrent neural networks to model human pose sequences, effectively integrating temporal information to reduce ambiguity in single-frame pose-conditioned approaches. This significantly enhances the capture of complex motions and non-rigid deformations, such as loose clothing dynamics. 
    
    \item We introduce a novel skeletal feature encoding method that leverages the regional independence of human motion. Instead of using entire poses as input, we partition the pose into multiple regions and encode them separately. Additionally, implicit encoding improves adaptability across different body parts, enhancing the generalization of clothing deformations.

    \item We present RealityAvatar, an efficient and accurate animatable digital human modeling framework leveraging 3D Gaussian Splatting to reconstruct high-fidelity digital humans from multi-view videos. Our method ensures stable geometry, precise appearance representation, and robust performance in handling complex poses and dynamic motions.

\end{itemize}

\section{Related Works}
\label{sec:related}

\myparagraph{Animatable Human Reconstruction.} The field of animatable human modeling has seen significant advancements through the use of neural implicit representations~\cite{chen2023fast, chen2021snarf, guo2023vid2avatar, kwon2021neural, kwon2023deliffas, xiu2022icon}, which aim to improve the quality of high-fidelity digital human reconstruction. Neural Radiance Fields (NeRF)~\cite{mildenhall2020nerf} and its variants learn a canonical neural radiance field and utilize inverse skinning or ray warping to model human shape and appearance. NeuralBody~\cite{peng2020neural} associates each SMPL vertex with a latent code to encode appearance and transforms it into observation space based on body pose, while ARAH~\cite{ARAH:ECCV:2022} adopts a neural signed distance function combined with SMPL shape priors to model human geometry, enabling detailed geometric representation in canonical space. Although these approaches achieve impressive results in free-view synthesis of static scenes, they still struggle to handle dynamic human motion. 

However, due to the difficulty of directly expressing clear surface structures using implicit 3D volumes, these methods often require large-scale multilayer perceptrons to fit human geometry and texture, resulting in high computational costs, prolonged training times, and slow inference speeds, making them impractical for many applications. In contrast, 3D Gaussian Splatting provides a more efficient and direct solution. By representing the surface of humans using 3D Gaussian, it offers a way to model dynamic human appearance more efficiently, with surface details directly encoded, enabling faster processing. D3GA~\cite{zielonka2023drivable} was the first to employ driveable 3D Gaussians and tetrahedral cages to create animatable human avatars, achieving promising results in geometry and appearance modeling. To improve rendering speed and high-resolution outputs on consumer-grade devices, GPS-Gaussian~\cite{zheng2024gps} proposed a Gaussian parameter map on sparse source views, jointly regressing Gaussian parameters with a depth estimation module without requiring fine-tuning or optimization. GaussianAvatar~\cite{hu2024gaussianavatar} integrated optimizable tensors with a dynamic appearance network to enhance motion capture, enabling real-time reconstruction and realistic novel-view animations of dynamic avatars. GVA~\cite{liu2024gva} further introduced a pose refinement technique by aligning normal maps and silhouettes to improve hand and foot pose accuracy. Additionally, to recover fine details on the human body, 3DGS-Avatar~\cite{qian20243dgs} replaced spherical harmonics (SH) with a shallow MLP for 3D Gaussian color modeling and applied geometric priors to regularize deformations, achieving realistic pose-dependent cloth deformations and effectively generalizing to novel poses. 

While these approaches have made remarkable progress in animatable human reconstruction, they still exhibit certain limitations. In particular, achieving both efficiency and high-quality human modeling remains challenging, especially when handling dynamic loose clothing. Existing methods either optimize efficiency at the expense of capturing fine-grained cloth dynamics or focus on detailed modeling while suffering from high computational costs and slow inference speeds. To better address these challenges, various strategies have been explored to improve the representation of clothing dynamics. In the following section, we review prior work on dynamic clothing modeling and evaluate their effectiveness in achieving both high fidelity and efficient motion representation for animatable humans.

\myparagraph{Dynamic Clothing Deformation in Human Motion.} In human motion modeling, incorporating temporal information is crucial for capturing dynamic appearance variations, particularly in clothing deformation. Early approaches predominantly relied on physics-based methods that simulate the physical properties of soft tissues to describe dynamic deformations. For example, ~\cite{larboulette2005dynamic} proposed a mechanical model using the Finite Element Method (FEM) to simulate the influence of skeletal acceleration on soft tissue motion. While this kind of method showed promise for bare-body modeling, it required manual parameter tuning for different materials, making it impractical for full-body clothed animation. Instead of modeling the underlying physics, subsequent research turned to data-driven approaches that learn deformation patterns directly from 4D scan data. Dyna~\cite{pons2015dyna} predicted soft tissue deformations directly from pose variations, but their focus on nude body geometry limited their applicability to clothed scenarios. DRAPE~\cite{guan2012drape} introduced motion history to simulate cloth draping and flow effects, achieving improved realism. However, its reliance on pre-defined templates restricted its ability to handle loose clothing under large motions.  

More recently, neural representation-based methods have been developed to improve generalization and enhance temporal modeling. RAM-Avatar~\cite{deng2024ram} introduced a dual-attention mechanism to address misalignment issues in per-frame neural texture synthesis, improving temporal coherence. While effective for tight clothing, it struggles with the complex deformations of loose garments. Dyco~\cite{chen2024within} proposed encoding incremental pose sequences (delta poses) and incorporating a local dynamic context encoder to mitigate inter-part dependencies, leading to improved accuracy in dynamic clothing reconstruction. However, despite these advancements, challenges remain in training efficiency and generalization to extreme motion scenarios.  

Despite notable progress in dynamic clothing modeling, existing methods still face key challenges in capturing the intricate deformations of loose garments. Even the sota method still face limitations in efficiency and adaptability to diverse motion patterns. These limitations highlight the need for a more effective framework that balances accuracy, efficiency, and temporal consistency in modeling loose clothing dynamics.

\section{Preliminaries}
\label{sec:prelim}
% In this section, we start by briefly reviewing the linear blend skinning (LBS) function for human articulation in ~\cref{sec:lbs}. We then explain 3D Gaussians Splatting in ~\cref{sec:3dgs}.

\myparagraph{SMPL and Linear Blend Skinning.}  A widely used approach for modeling human articulation is to employ a parametric human body model, such as SMPL~\cite{SMPL:2015}, which defines human geometry in a canonical space~\cite{guo2023vid2avatar,jiang2023instantavatar,jiang2022neuman,li2022tava,peng2022animatable,ARAH:ECCV:2022,weng2022humannerf}. To animate the body under arbitrary poses, Linear Blend Skinning (LBS)~\cite{anguelov2005scape,SMPL:2015,SMPL-X:2019,osman2020star} is commonly used, where each point on the body surface undergoes a weighted transformation based on the influence of nearby skeletal bones. Given a point $\mathbf{x}_c$ in canonical space, its transformed position $\mathbf{x}_o$ in the posed observation space is computed as a weighted sum of bone transformations:

\begin{equation}
    \mathbf{x}_o = \sum_{b=1}^{B} w_b \mathbf{B}_b \mathbf{x}_c
\end{equation}
where $B$ represents the total number of skeletal bones ($B=24$ in the SMPL model), $\mathbf{B}_b$ is a $4 \times 4$ transformation matrix representing the rotation and translation of bone $b$, and $w_b$ is the corresponding skinning weight, which satisfies:

\begin{equation}
    \sum_{b=1}^{B} w_b = 1, \quad w_b \in [0,1]
\end{equation}
These skinning weights determine the extent to which each bone influences the deformation of a given point. Traditionally, they are precomputed based on the body mesh topology, but modern approaches often learn them dynamically using neural networks.  

LBS provides a computationally efficient and differentiable way to model human motion, making it well-suited for applications such as animation and motion capture. However, it has limitations in handling complex non-rigid deformations, such as soft-tissue dynamics and loose clothing motion, since it assumes a purely skeletal-driven transformation. Alternative methods, such as inverse skinning, attempt to establish correspondences between the observation and canonical spaces but often introduce ambiguities. More recent works explore learning-based techniques to improve skinning weight prediction, allowing for more accurate deformation modeling in data-driven human reconstruction.

\myparagraph{3D Gaussian Splatting.} 3D Gaussian Splatting~\cite{kerbl20233d} has emerged as an efficient representation for 3D scene modeling and rendering. Unlike volumetric methods, which represent a scene as a dense field requiring costly volumetric sampling, 3DGS models a scene as a collection of discrete 3D Gaussian primitives. Each 3D Gaussian is parameterized by its mean position $\mathbf{x}$, covariance matrix $\boldsymbol{\Sigma}$, opacity $\alpha$ and view-dependent color represented by spherical harmonics coefficients. During rendering, these 3D Gaussians are projected onto the image plane, where their contributions are blended based on depth ordering via alpha compositing. To ensure positive semi-definiteness, the covariance matrix is represented by a scaling matrix $\mathbf{S}$ and rotation matrix $\mathbf{R}$. The covariance is factorized as:

\begin{equation}
    \boldsymbol{\Sigma} = \mathbf{R} \mathbf{S} \mathbf{S}^T \mathbf{R}^T,
\end{equation}

where $\mathbf{R}$ is a rotation matrix derived from a unit quaternion, and $\mathbf{S}$ is a diagonal matrix that represents anisotropic scaling along the principal axes of the Gaussian. This decomposition allows each Gaussian to deform anisotropically while remaining differentiable during optimization. To render a 3D scene, each Gaussian is projected onto the 2D image plane, where it appears as an elliptical splat. The projected covariance $\boldsymbol{\Sigma}'$ is computed using the camera projection matrix $\mathbf{W}$ and the Jacobian $\mathbf{J}$ of the perspective transformation:

\begin{equation}
    \boldsymbol{\Sigma}' = \mathbf{J} \mathbf{W} \boldsymbol{\Sigma} \mathbf{W}^T \mathbf{J}^T.
\end{equation}
This determines the elliptical footprint of the Gaussian in image space, controlling its blending behavior. Rendering is performed via differentiable rasterization, where overlapping Gaussians contribute to each pixel using alpha blending. The final pixel color $C$ is computed as:

\begin{equation}
    C = \sum_{i} \alpha_i w_i c_i, \quad \text{where} \quad w_i = \prod_{j=1}^{i-1} (1 - \alpha_j),
\end{equation}

where $\alpha_i$ represents the learned opacity of the $i$-th Gaussian weighted by its projected probability density, and $c_i$ is the view-dependent color modulated by spherical harmonics.

Compared to volumetric representations such as NeRF, 3DGS provides a more memory-efficient and computationally efficient solution while maintaining high rendering quality. Recent advancements~\cite{wu20234dgaussians,zhou2024drivinggaussian,zhou2024hugs} have focused on optimizing Gaussian parameters for dynamic scenes, as well as introducing adaptive pruning and refinement strategies to balance efficiency and accuracy. These improvements highlight the potential of 3DGS as a powerful framework for real-time 3D modeling and rendering applications.

\begin{figure*}
\centering
\includegraphics[width=1.\textwidth]{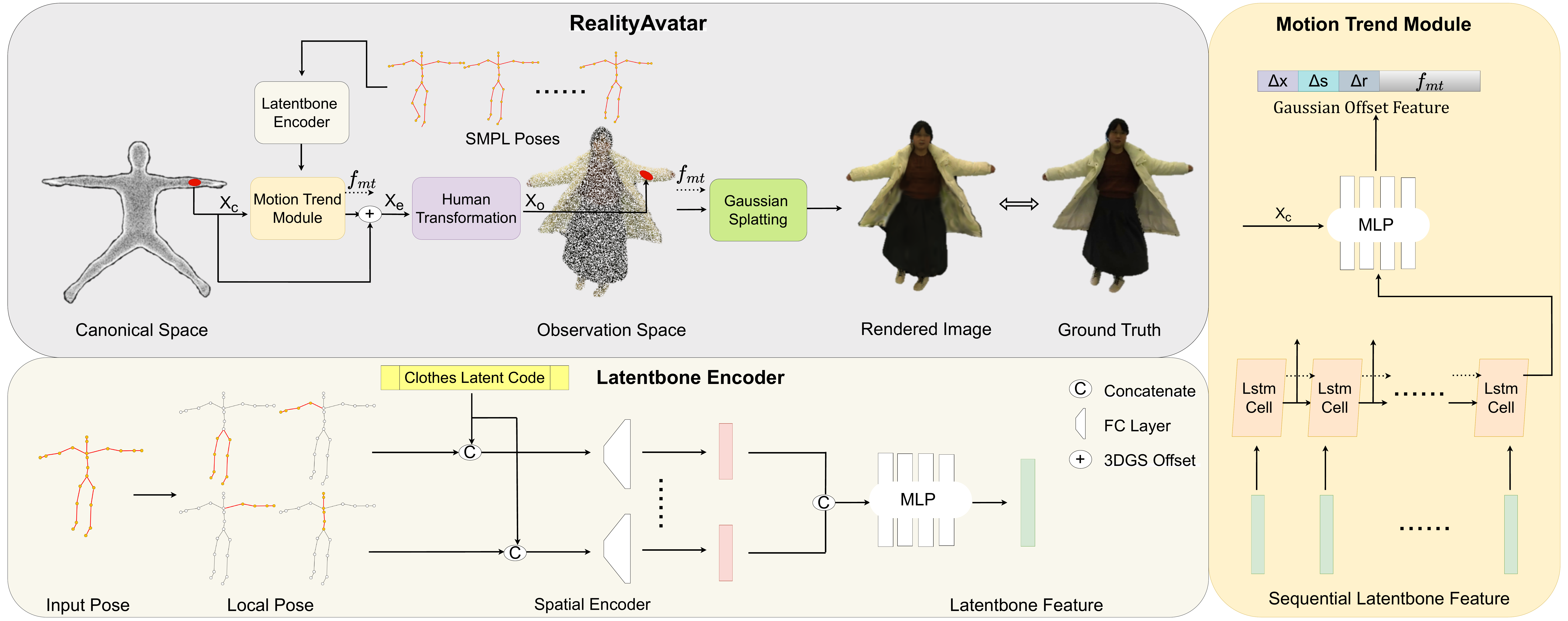}
\caption{The overall pipeline of our method. The human body is initialized in canonical space based on the SMPL model. Using the motion trend module and human transformation, we deform the 3D Gaussian representation from canonical space to observation space. To extract meaningful pose features, we introduce the latentbone encoder, which captures skeletal representations that influence clothing dynamics. In the motion trend module, a recurrent neural network is employed to encode temporal dependencies, modeling different dynamic effects of clothing under similar poses. These components work together to achieve realistic motion-driven human reconstruction.}
\label{fig:overview}
\end{figure*}

\section{Methods}
\label{sec:methods}
We illustrate the pipeline of our method in \cref{fig:overview}. The input to our method consists of a multi-view video captured with a calibrated camera, fitted SMPL parameters, and foreground masks. Our approach optimizes a set of 3D Gaussians in canonical space, which is then deformed to the observation space and rendered from the given camera. For a set of 3D Gaussians $\{\mathcal{G}^{(i)}\}_{i=1}^N$, we store the following properties at each point: position $\mathbf{x}$, scaling factor $\mathbf{s}$, rotation quaternion $\mathbf{q}$, opacity $\boldsymbol{\alpha}$, and a color feature vector $\mathbf{c}$. We initialize our canonical 3D Gaussians $\{G_c\}$ by randomly sampling $N = 35k$ points on the SMPL mesh surface under a canonical pose.  

We propose to decompose the transformation from canonical space to observation space into two components: a motion trend module, which captures the overall motion trend, and a human transformation module, which characterizes motion-driven shape variations and completes the mapping from canonical space to observation space. Furthermore, to improve the generalization ability of our model, we design a latentbone encoder to enhance the modeling of pose-driven clothing deformation.    

\subsection{Latentbone Encoder}
\label{sec:latent_bone}
In the process of motion, we define a state where a particular posture is maintained for a certain period before reaching a stable motion condition as the Post-Motion Steady State. This term differentiates it from a completely static state by emphasizing that the human body was in motion before reaching this condition. It is evident that even under the same posture, clothing deformations can vary significantly. To simplify this complex problem, we assume that the clothing deformation in the Post-Motion Steady State primarily depends on the current posture parameters. Therefore, we attempt to model clothing deformation in this transitional state using the latentbone encoder. Building upon the idea that clothing deformation may be influenced by a latent skeletal structure, as suggested in ~\cite{lei2023gart}, we hypothesize that it is primarily determined by the human skeleton and the inherent shape characteristics of the clothing.

On the one hand, previous studies~\cite{chen2024within,deng2024ram} have shown that modeling clothing deformation using local pose parameters instead of global posture parameters helps reduce artifacts in non-sampled regions and improves the overall quality of clothing reconstruction. Inspired by this, we divide the human body into four parts and similarly partition the corresponding SMPL pose parameters into four groups: left arm $\mathbf{p}_{lm}$, right arm $\mathbf{p}_{rm}$, legs $\mathbf{p}_{lg}$ and torso $\mathbf{p}_{to} $. Notably, we observe that clothing deformation in the leg region is influenced not only by the motion of an individual leg but also by the relative rotation between the two legs. This effect is particularly pronounced when reconstructing loose-fitting clothing.

On the other hand, we introduce a randomly initialized latent code, termed the clothes latent code $\mathcal{Z}_c$, to assist the network in learning. This variable is designed to represent the degree of cloth fluttering under the same motion posture but with different clothing. Our experimental results confirm that incorporating this latent code is both meaningful and effective. For each body part, we concatenate its extracted features with the clothes latent code $\mathcal{Z}_c$, then pass them through separate single-layer fully connected networks. The outputs are then concatenated and fed into a shallow MLP, which produces the encoded feature representation for a single frame in the motion sequence.
Our latentbone encoder is formulated as follows: 
\begin{equation}
    \{f_p\} = \mathcal{F}_{\theta_{le}} 
    \left( \{ \mathbf{p}_{lm},\mathbf{p}_{rm}, \mathbf{p}_{lg}, \mathbf{p}_{to} \} ; \mathcal{Z}_c \right)
    \label{eq:latentbone_encoder}
\end{equation}
$\theta_{le}$ represents the learnable parameters of the latentbone encoder. Since our method does not rely solely on the pose parameters of the current frame, we use  
$\{ \mathbf{p}_{lm},\mathbf{p}_{rm}, \mathbf{p}_{lg}, \mathbf{p}_{to} \}$ to represent the set of pose parameters across all frames. The encoded feature obtained from the latentbone encoder is denoted as \(f_p \). 

\subsection{Motion Trend Module}
\label{sec:motion_trend}
The above method is insufficient to model realistic clothing motion. In real-world scenarios, the human body rarely remains in a post-motion steady state; instead, motion is typically continuous and dynamic. In particular, modeling the transition from continuous movement to a stationary state presents a significant challenge. While clothing deformations during motion can be partially inferred from body pose parameters, the transition phase is more complex. During this period, body pose parameters remain unchanged, yet clothing continues to exhibit dynamic behavior rather than instantly settling into a static state.  
As a result, the clothing state in the current frame is influenced not only by the present body posture but also by previous motion states. To address this, we incorporate a temporal modeling network.  

When selecting a suitable temporal network for modeling clothing motion, we evaluated multiple architectures. Since clothing dynamics exhibit strong temporal dependencies, we require a model capable of capturing long-term temporal relationships.

Given the relatively limited scale of our dataset and the strong local temporal correlations in clothing dynamics, the benefits of attention-based mechanisms in this task are relatively constrained. However, recurrent neural networks and their variants, such as long short-term memory (LSTM) networks~\cite{yu2019review,dipietro2020deep}, have demonstrated robust performance in sequence modeling, particularly in tasks with strong temporal dependencies. Based on these considerations, we conducted comparative experiments on different temporal modeling approaches. Under the same experimental settings, we found that LSTM achieved superior accuracy and stability in predicting clothing dynamics, effectively capturing fine-grained deformations and reducing cumulative errors.

Therefore, we ultimately chose LSTM as the core component for temporal modeling, as it enhances the network’s ability to predict clothing motion deformation, with our experimental results validating its effectiveness. We formulate the motion trend module as:
\begin{equation}
    \{\mathcal{G}_e\}=\mathcal{F}_{\theta_{mt}}\left(\{\mathcal{G}_c\}; \{f_p\}\right)
    \label{eq:mt_gaussian_set}
\end{equation}

where \(\{\mathcal{G}_e\}\) represents the 3D Gaussians deformed by the motion trend module, and $\{f_p\}$ denotes the sequential features encoded by the latent bone encoder. Given a sequence step size \(s\) and a sequence length \(t\), the feature sequence is defined as:  
\begin{equation}
    \{f_p\} = f_{p-(t-1)s}, ..., f_{p-s}, f_{p} 
\end{equation}
To capture temporal dependencies, we encode the feature sequence using an LSTM, enabling the current frame’s output features to incorporate the influence of previous frames. We take the output of the final LSTM cell at the last time step, which represents the current frame, as the input to a three-layer MLP decoder. Meanwhile, the decoder takes the canonical position \(\mathbf{x}_c\) as input and predicts the offsets for the Gaussian’s position$\Delta\mathbf{x}$, scale$\Delta\mathbf{s}$, and rotation$\Delta \mathbf{q}$, along with a feature vector \({f}_{mt}\). The canonical Gaussian is deformed as follows:  
\begin{align}
    \mathbf{x}_e &= \mathbf{x}_c + \Delta\mathbf{x} \label{eq:mt_deform_xyz}\\
    \mathbf{R}_e &= \Delta\mathbf{R} \cdot \mathbf{R}_c \label{eq:mt_deform_rot}\\
    \mathbf{s}_e &= \mathbf{s}_c \cdot \exp(\Delta\mathbf{s}) \label{eq:mt_deform_scale}
\end{align}
where \(\mathbf{R}_c\) and \(\Delta \mathbf{R}\) are the rotation matrix representations of the canonical Gaussian quaternion \(\mathbf{q}_c\) and the predicted rotation offset \(\Delta \mathbf{q}\).

\subsection{Human Transformation}
\label{sec:human_transf}
We subsequently map the motion trend module deformed 3D Gaussians \(\{\mathcal{G}_e\}\) into the observation space using a transformation step similar to the rigid transformation module in~\cite{qian20243dgs}. This module applies a linear blend skinning approach to ensure articulated motion consistency. Specifically, a skinning MLP \( \mathcal{F}_{\theta_h} \) is trained to predict skinning weights at the Gaussian center position \(\mathbf{x}_e\), which are then used to apply the following transformation:
\begin{align}
    \mathbf{T} &= \sum\nolimits_{b=1}^B \mathcal{F}_{\theta_h}(\mathbf{x}_e)_b\mathbf{B}_b \label{eq:r_gaussian_transform} \\
    \mathbf{R}_o &= \mathbf{T}_{1:3,1:3} \mathbf{R}_e \label{eq:r_gaussian_rot} \\
    \mathbf{x}_o &= \mathbf{T} \mathbf{x}_e \label{eq:r_gaussian_xyz} 
\end{align}

\subsection{Guassian Splatting}
\label{sec:splatting}
In the Gaussian splatting stage, we adopt the differentiable rendering pipeline commonly used in 3DGS, as it has been demonstrated to be both efficient and effective. However, unlike conventional methods that solely rely on learnable spherical harmonics coefficients \(\bm{\gamma}(\mathbf{d})\) and precomputed basis functions to model view-dependent color, our approach incorporates additional factors to improve realism.  

The rendered pixel color of a clothed human avatar is influenced not only by the stored spherical harmonics representation but also by local surface deformations. Fine wrinkles in clothing, for instance, can introduce self-occlusions that significantly affect shading. To better capture these variations, similar to~\cite{qian20243dgs}, we utilize both the learned SH coefficients and the basis functions computed from the canonicalized viewing direction \(\bm{\gamma}(\hat{\mathbf{d}})\). Additionally, a 16-dimensional feature vector \(f_{mt}\) generated by the motion trend module is incorporated to account for dynamic surface changes.  Furthermore, to mitigate lighting inconsistencies caused by the subject’s global motion, a per-frame latent code \(\mathcal{Z}_r\) is introduced, similar to prior work. A neural network parameterized by \(\theta_c\), representing the learnable weights of the model, takes these components as input and predicts the final view-dependent appearance. Empirically, a compact MLP with a single hidden layer of 64 neurons is found to be sufficient to capture fine appearance details while maintaining computational efficiency.

\subsection{Optimization}
\label{sec:opt}
We jointly optimize the canonical 3D Gaussians ${\mathcal{G}c}$ and the parameters $\theta_{le}$, $\theta_{mt}$, $\theta_h$, and $\theta_c$ of the latentbone encoder network, the motion trend module network, the skinning network, and the color network, respectively. To train the rendering framework, we adopt the common L1 loss, mask loss, and perceptual loss using a VGG19 network. The total loss is defined as a weighted sum of these components:

\begin{equation}
    \mathcal{L}_{total} = \mathcal{L}_{L1} + \lambda_1 \mathcal{L}_{mask} + \lambda_2 \mathcal{L}_{percep} + \lambda_3 \mathcal{L}_{skin}
\end{equation}

To further improve the accuracy of human transformation, inspired by ~\cite{ARAH:ECCV:2022}, we utilize a skinning loss to constrain the learned skinning weights. Similar to the segmentation strategy used in the latentbone encoder, we sample points from different body parts of the SMPL model and enforce consistency between the predicted skinning weights and the interpolated SMPL skinning weights. Specifically, given a set of sampled points $\mathbf{X}$ on the surface of the canonical SMPL mesh, we regularize the skinning network by minimizing the difference between the predicted weights $\mathcal{F}_{\theta_h}(\mathbf{x})$ and the ground-truth skinning weights $\mathbf{w}$ interpolated with barycentric coordinates:

\begin{equation}
    \mathcal{L}_{skin} = \frac{1}{|\mathbf{X}|} \sum_{\mathbf{x} \in \mathbf{X}} \left\| \mathcal{F}_{\theta_h}(\mathbf{x}) - \mathbf{w} \right\|^2.
\end{equation}

This loss encourages the skinning network to learn anatomically consistent deformations, improving the stability and realism of motion-driven shape transformations.

\begin{figure*}[t]
    \centering
    \hspace{-7mm}
    \includegraphics[width=1.05\textwidth]{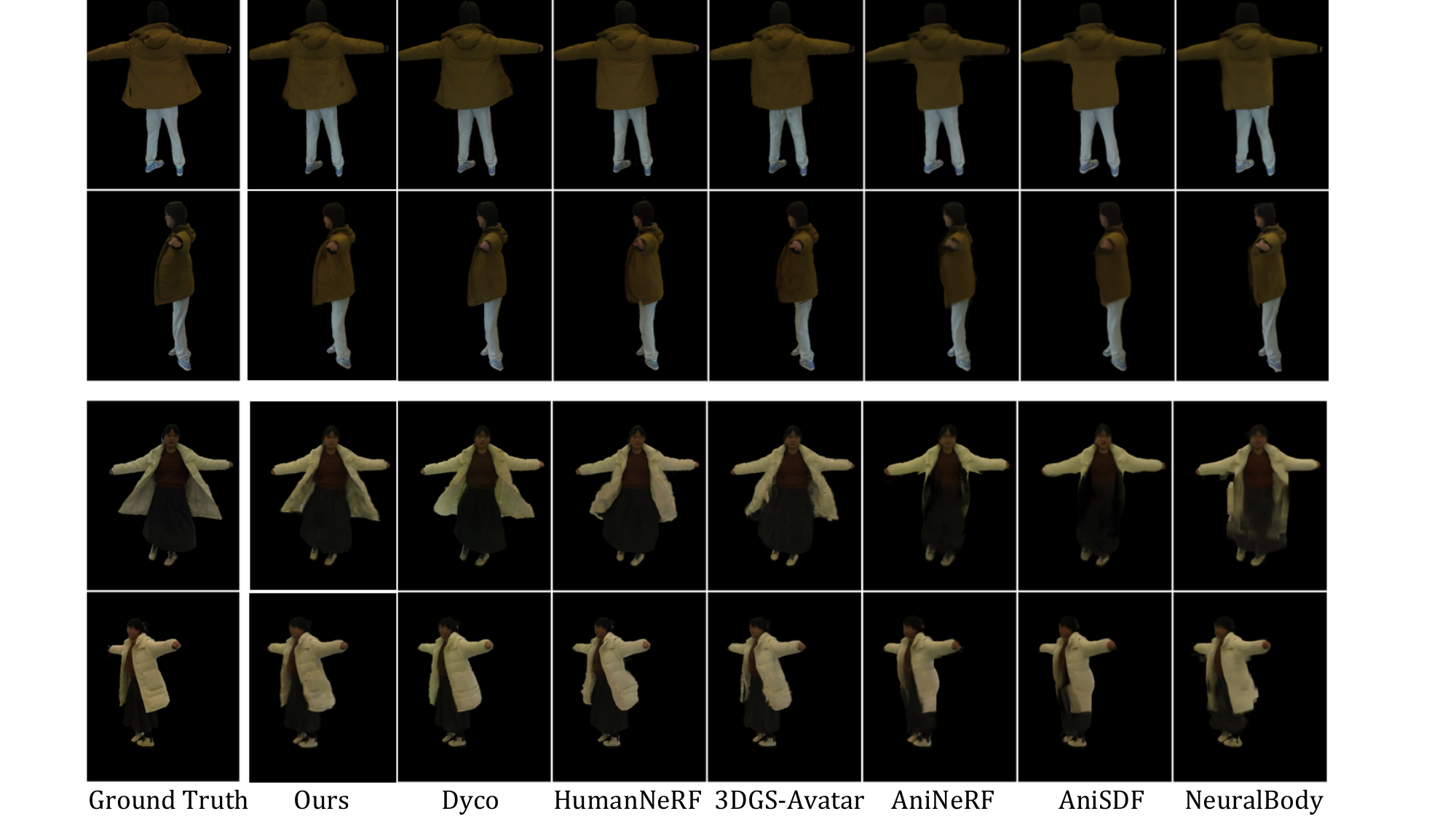}
    \vspace{5pt} % 增加两张图片间的间距
    \includegraphics[width=1\textwidth]{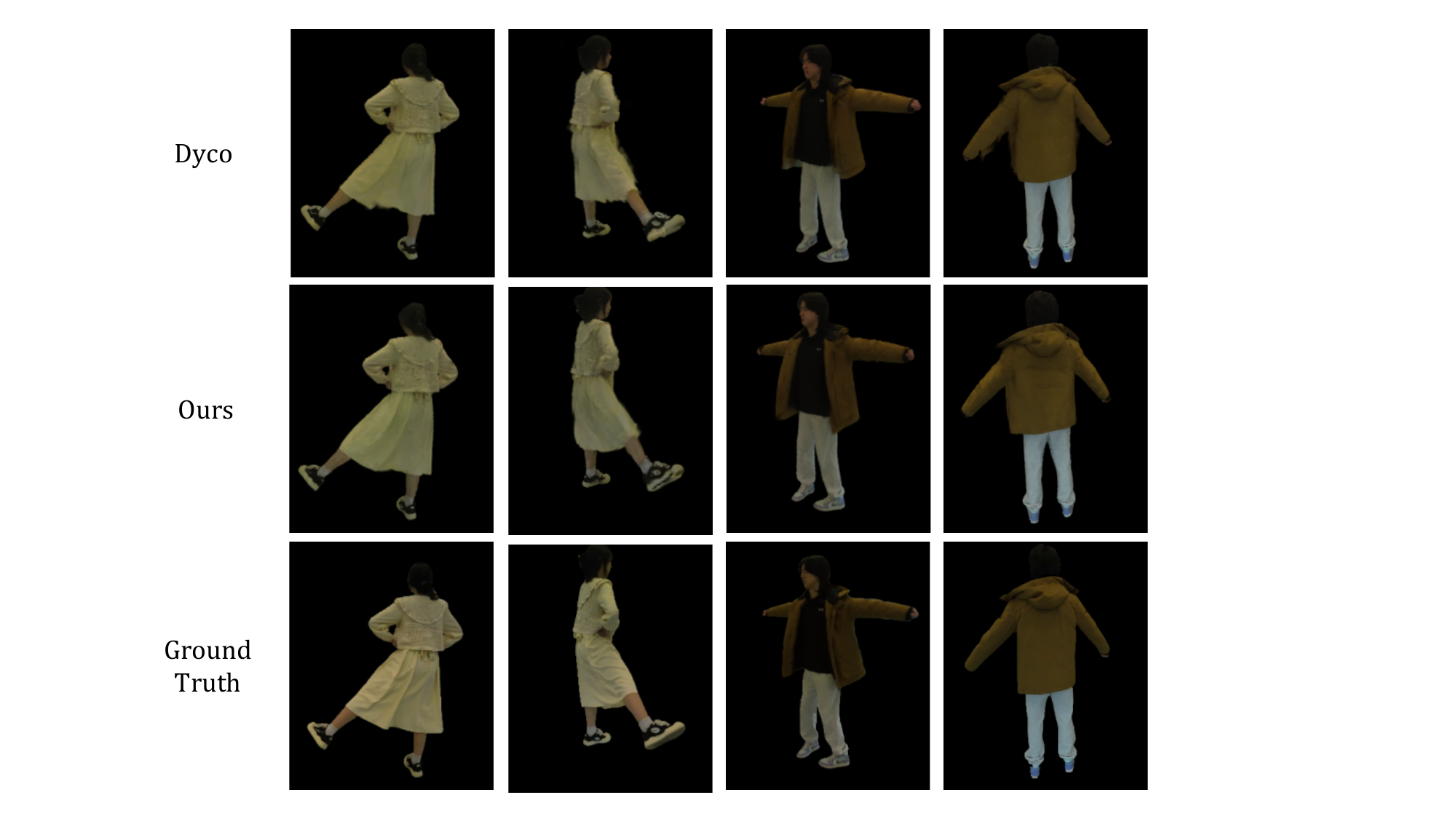}
    \caption{Qualitative comparison on the I3D-Human dataset. Top: Novel View Comparison. Bottom: Novel Pose Comparison. Note that in terms of novel view generation, some baseline generation graphs are quoted from Dyco~\cite{chen2024within}. For novel pose comparison, we focus on comparing with the Dyco~\cite{chen2024within}, as this method also specializes in loose clothing human body modeling and driving.}
    \label{fig:qualitative_comparison}
\end{figure*}

\begin{table*}[t]
\centering
\setlength{\fboxsep}{0pt}
\caption{Quantitative evaluation on I3D-Human dataset~\cite{chen2024within}.  showing average scores across four sequences. \textbf{Bold} indicates best performance, while \underline{underlined} denotes second best results.}
\label{tab:compare_I3D}
\setlength{\tabcolsep}{4pt} 
\renewcommand{\arraystretch}{1}
\begin{tabular}{l|c|ccc|ccc}
    \toprule
    \multirow{2}{*}{Methods} & \multirow{2}{*}{Training time} & \multicolumn{3}{c|}{Novel View} & \multicolumn{3}{c}{Novel Pose} \\
    \cline{3-8}
    & & PSNR↑ & SSIM↑ & LPIPS↓ & PSNR↑ & SSIM↑ & LPIPS↓ \\
    \hline
    NeuralBody~\cite{peng2020neural} & $\sim$12h & 30.33 & 0.9681 & 60.89 & 28.80 & 0.9604 & 67.59 \\
    AniNeRF~\cite{zhou2024animatable} & $\sim$30h & 29.29 & 0.9662 & 61.95 & 28.48 & 0.9628 & 64.85 \\
    AniSDF~\cite{zhou2024animatable} & $\sim$30h & 29.20 & 0.9670 & 58.94 & 28.34 & 0.9632 & 62.18 \\
    HumanNeRF~\cite{weng2022humannerf} & $\sim$10h & 29.53 & 0.9678 & 42.04 & 28.78 & 0.9644 & 45.70 \\
    3DGS-Avatar~\cite{qian20243dgs} & \textbf{$\sim$0.5h} & 30.62 & 0.9712 & 39.74 & 29.21 & 0.9658 & 44.61 \\
    Dyco~\cite{chen2024within} & $\sim$12h & \underline{31.22} & \underline{0.9738} & \textbf{34.54} & \textbf{30.12} & \textbf{0.9691} & \textbf{39.55} \\
    \hline
    Ours & \underline{$\sim$0.6h} & \textbf{31.87} & \textbf{0.9752} & \underline{34.78} & \underline{30.10} & \underline{0.9689} & \underline{41.31} \\
    \bottomrule
\end{tabular}

\end{table*}

\section{Experiments}
\label{sec:exp}
In this section, we evaluate the proposed approach on standard benchmarks and compare it with recent state-of-the-art methods. We assess reconstruction quality using PSNR, SSIM, and LPIPS~\cite{zhang2018unreasonable} as evaluation metrics. Note that LPIPS in all the tables are scaled up by $1000$. Since our method primarily focuses on improving multi-view reconstruction for subjects with loose clothing, we follow common practices by utilizing multi-view data for training.  

To ensure a comprehensive evaluation, we conduct experiments across various datasets, covering different types of clothing and motion sequences. We also analyze the impact of model components through a systematic ablation study, providing insights into their role in improving reconstruction fidelity. To further investigate the impact of different components in our model, we conduct a systematic ablation study, providing insights into their role in rendering quality.

\subsection{Evaluation Dataset}
\label{sec:dataset}
\myparagraph{I3D-Human~\cite{chen2024within}.} The I3D-Human dataset captures variations in clothing appearance while maintaining nearly identical poses. Unlike existing benchmarks, it includes subjects wearing loose garments such as dresses and lightweight jackets, performing dynamic movements with acceleration and deceleration. These motions involve abrupt stops after spinning, swaying, and sleeve flapping, making the dataset well-suited for evaluating clothing dynamics. We use I3D-Human as the primary benchmark for quantitative evaluation, selecting four sequences (ID$1$\_$1$, ID$1$\_$2$, ID$2$\_$1$, ID$3$\_$1$) and following its standard training/test split.

\myparagraph{ZJU-Mocap~\cite{peng2020neural}.} While our primary experiments are conducted on a dataset featuring loose clothing, we additionally incorporate the ZJU-MoCap dataset to further evaluate the generalization ability of our method. We use nine complete sequences from ZJU-MoCap, following the training/test split defined in NeuralBody~\cite{peng2020neural}. Since this dataset primarily consists of subjects performing repetitive motions with limited pose variations, it is less suitable for assessing novel pose synthesis. Instead, we leverage ZJU-MoCap to supplement our evaluation on novel view synthesis, ensuring a more comprehensive comparison with existing methods.

\subsection{Comparison with Baselines}
\label{sec:comparison}
We compare our method with NeuralBody~\cite{peng2020neural}, HumanNeRF~\cite{weng2022humannerf}, 3DGS-Avatar~\cite{qian20243dgs}, AniNeRF~\cite{zhou2024animatable}, AniSDF~\cite{zhou2024animatable}, and Dyco~\cite{chen2024within} on the I3D-Human dataset under multi-view training. Since I3D-Human primarily consists of subjects wearing loose clothing, it serves as the main benchmark for evaluating our method. To further assess the generalization ability of our approach, we also conduct experiments on the ZJU-MoCap dataset.

\begin{table}[ht]
\centering
\setlength{\fboxsep}{0pt}
\caption{Quantitative Comparison on ZJU-Mocap~\cite{peng2020neural} for Novel View Synthesis. \textbf{Bold} indicates best performance, while \underline{underlined} denotes second best results.}
\label{tab:compare_ZJU}
\renewcommand{\arraystretch}{1}
\begin{tabular}{l|ccc}
    \toprule
    Methods & PSNR↑ & SSIM↑ & LPIPS↓\\
    \hline
    AniNeRF~\cite{zhou2024animatable} & 32.28 & 0.9750 & 47.53 \\
    AniSDF~\cite{zhou2024animatable} & 32.16 & 0.9767 & 40.63 \\
    HumanNeRF~\cite{weng2022humannerf} & 32.13 & 0.9783 & 22.30 \\
    3DGS-Avatar~\cite{qian20243dgs} & \underline{32.81} & 0.9791 & 23.76 \\
    Dyco~\cite{chen2024within} & 32.80 & \underline{0.9800} & \textbf{20.32} \\
    \hline
    Ours & \textbf{33.20} & \textbf{0.9804} & \underline{22.27} \\
    \bottomrule
\end{tabular}
 % 结束 \resizebox
\end{table}

\begin{figure*}
    \centering
    \includegraphics[width=\textwidth]{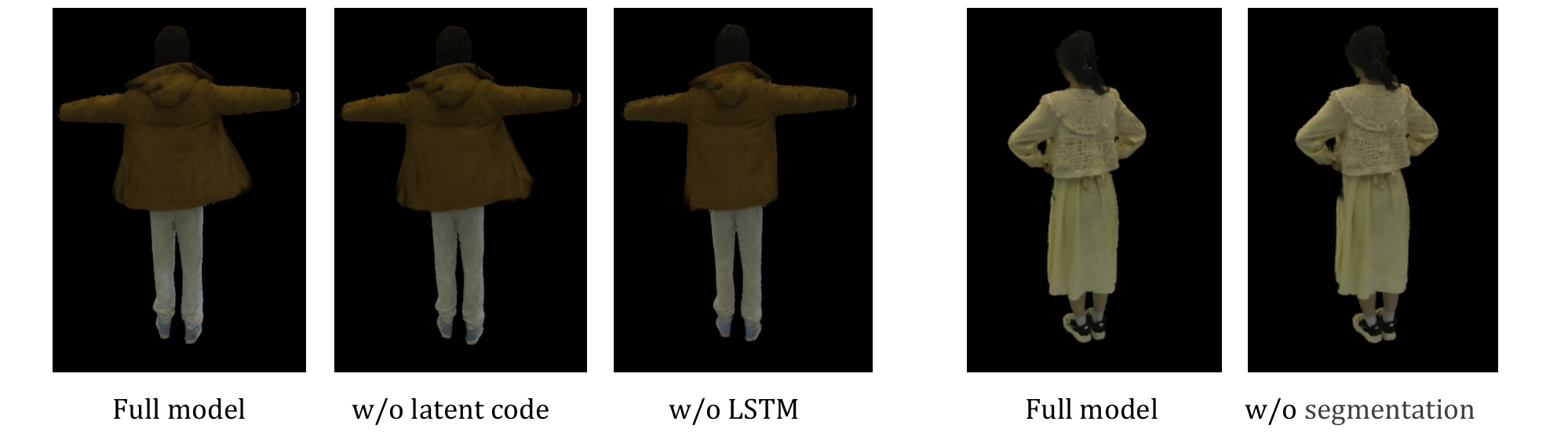}
    \caption{Ablation Study on I3D-Human Dataset. }
    \label{fig:ablation}
\end{figure*}
The quantitative results, summarized in Table~\ref{tab:compare_I3D}, demonstrate that our method achieves state-of-the-art performance in novel view synthesis, obtaining the highest PSNR and SSIM scores. For novel pose animation, our method achieves the highest performance in terms of PSNR and SSIM, surpassing existing baselines in objective metrics. Our approach consistently preserves fine geometric details and reconstructs sharper images, highlighting its advantages in structural fidelity and perceptual quality. To supplement our evaluation, we further validate our method on the ZJU-MoCap dataset, which primarily consists of subjects in tight clothing performing repetitive motions. The results, presented in Table~\ref{tab:compare_ZJU}, indicate that our approach achieves competitive performance in novel view synthesis, ranking second in PSNR while attaining the highest SSIM.  

Regarding LPIPS, while our method performs competitively and surpasses most baselines, Dyco achieves a slightly lower LPIPS score. Given that LPIPS is sensitive to texture variations and global illumination effects, this result suggests that Dyco may better preserve certain high-frequency details. Nevertheless, our approach achieves a strong balance between perceptual quality and structural consistency, ensuring more stable and realistic reconstructions.

\subsection{Ablation Study}
\label{sec:ablation}
This section examines the influence of LSTM network in the motion rend module, the latent code $\mathcal{Z}_c$ and the part segmentation of the SMLP pose parameters in latentbone encoder. Table~\ref{tab:ablation} presents the quantitative results from our ablation study on the I3D-Human dataset, evaluating the impact of each component on novel view synthesis performance. Figure~\ref{fig:ablation} provides a visual comparison of the effects of these components on the model’s output.
\begin{table}[h]
\caption{Ablation Study on I3D-Human~\cite{peng2020neural}. Quantitative evaluation of the impact of different components on novel view synthesis performance.}
 \label{tab:ablation}
 \centering
 \begin{tabular}{@{}lccc}
 \toprule
 Methods         
 & PSNR$\uparrow$
 & SSIM$\uparrow$
 & LPIPS$\downarrow$\\ \hline
 Full model               
 & \textbf{31.87}
 & \textbf{0.9752}
 & \textbf{34.78}
\\
  w/o latent code $\mathcal{Z}_c$
 & 31.81
 & \textbf{0.9752}
 & 35.41
\\
  w/o LSTM      
 & 30.88
 & 0.9721
 & 38.04
 \\
 w/o part segmentation  
 & 31.61
 & 0.9746
 & 35.78
  \\
 \bottomrule
 \end{tabular}
\end{table}

The absence of the LSTM network leads to a noticeable drop in all metrics, particularly a significant decrease in PSNR and an increase in LPIPS. This highlights the importance of temporal modeling in capturing motion-dependent clothing dynamics. Without LSTM, the model struggles to account for temporal consistency, resulting in reduced reconstruction fidelity. Disabling part segmentation in the latentbone encoder also degrades performance, albeit to a lesser extent. The LPIPS score increases while PSNR and SSIM show minor reductions, suggesting that part-wise pose encoding contributes to refining local details and reducing artifacts. This confirms that modeling local pose variations independently improves the overall reconstruction quality. These findings demonstrate that each component plays a crucial role in enhancing both structural fidelity and perceptual quality, validating the effectiveness of our design choices.

\section{Conclusion}
\label{sec:conclusion}

In this paper, we propose \textbf{RealityAvatar}, a novel framework for high-fidelity animatable digital human modeling based on 3D Gaussian Splatting. Our method effectively captures complex clothing deformations and motion dynamics while maintaining geometric consistency. By incorporating a motion trend module and a latentbone encoder, we introduce a structured approach to modeling pose-dependent deformations and temporal variations in clothing behavior. Extensive experiments on standard benchmarks demonstrate the effectiveness of our method in both novel view synthesis and novel pose animation. Our results highlight improvements in structural fidelity and perceptual quality, showcasing its ability to generate realistic and dynamically consistent digital humans.

While our approach achieves strong performance, there is still room for improvement in terms of training efficiency and fine-grained texture reconstruction. Future work will explore optimizing training speed while maintaining high-quality results and enhancing texture detail preservation for more realistic rendering. We believe our framework provides a solid foundation for future research in 3D Gaussian-based animatable digital human reconstruction.

{
    \small
    \bibliographystyle{ieeenat_fullname}
    \bibliography{main}
}

\end{document}